\documentclass[pdflatex,sn-mathphys-num]{sn-jnl}

\usepackage{graphicx}%
\usepackage{multirow}%
\usepackage{amsmath,amssymb,amsfonts}%
\usepackage{amsthm}%
\usepackage{mathrsfs}%
\usepackage[title]{appendix}%
\usepackage{xcolor}%
\usepackage{textcomp}%
\usepackage{manyfoot}%
\usepackage{booktabs}%
\usepackage{algorithm}%
\usepackage{algorithmicx}%
\usepackage{algpseudocode}%
\usepackage{listings}%
\usepackage{tabularx}

\theoremstyle{thmstyleone}%
\theoremstyle{thmstyletwo}%

\theoremstyle{thmstylethree}%

\begin{document}

\title[Article Title]{Hybrid NARX-LLM for Greenland Iceberg Discharge: Prompt-Driven Residual Correction}


\author[1,2,3,*]{\fnm{Yiquan} \sur{Gao}}

\author[2]{\fnm{Duohui} \sur{Xu}}

\affil[1]{\orgname{Heriot-Watt University}}

\affil[2]{\orgname{StudioYG}}


\abstract{
Greenland iceberg discharge exhibits complex nonlinear dynamics with limited observability, challenging traditional predictive models. We present a Hybrid NARX--LLM framework that combines a nonlinear autoregressive model with exogenous inputs (NARX) and a large language model (LLM) for residual correction. We further propose a Physics-Informed Prompt (PIP) method that transforms unstructured physical knowledge into structured prompts for zero-shot in-context reasoning. The primary objective is to explore the corrective potential of this framework for modeling Greenland iceberg discharge, rather than merely optimizing predictive accuracy. The NARX component captures intrinsic temporal dependencies, while the LLM, guided by PIP, encodes glacier dynamics and environmental drivers and perceives key trend patterns to correct systematic prediction errors. This integration allows the model to reason about unmodeled factors and produce interpretable residuals, enhancing overall predictive accuracy. Applied to Greenland iceberg discharge time series, our approach addresses extreme events that are difficult to predict due to rare variations and nonstationary trends, a limitation often overlooked by traditional methods. By fusing structured time-series modeling with knowledge-driven foundation AI, the framework offers a scalable and interpretable pathway to bridge data-limited climate forecasting with physics-informed LLM reasoning. The code is available. 
}

\keywords{NARX, large language models, hybrid machine learning, Greenland ice sheet, iceberg calving, climate modeling, physics-informed learning.}

\makeatletter
\Footnotetext{}{\kern-1em * First and Corresponding author.}
\Footnotetext{}{\kern-1em \textsuperscript{3} This work was completed before the first author's affiliation with Heriot-Watt University and was conducted entirely independently of the institution.}
\makeatother


\maketitle

\section{Introduction}
Greenland ice sheet mass loss has become one of the dominant contributors to global sea level rise in recent decades~\cite{ruane2024synthesis}. Among the various mass-loss pathways, iceberg discharge (calving flux) plays a critical role in the dynamics of marine-terminating glaciers. Accurate prediction of iceberg discharge is therefore essential for improving sea level rise projections and understanding ice--ocean interactions in a warming climate~\cite{ding2021increasing}.

However, iceberg discharge is governed by nonlinear and partially observed ice--ocean processes, resulting in strong non-stationarity and limited predictability~\cite{fitzmaurice2017nonlinear,benn2017glacier}. This makes it difficult for data-driven models to generalize across different dynamical regimes.

Widely adapted as a benchmark for iceberg discharge forecasting, NARX models can capture nonlinear temporal dependencies between exogenous inputs and discharge time series~\cite{diaconescu2008use,kelley2024comparison}, but they remain sensitive to distribution shifts and prone to inherent prediction delays. Physics-based models~\cite{van2002calving} are often limited by simplified representations of calving dynamics. Although increasing data availability can partially alleviate some generalization limitations of NARX and other data-driven models, it suffers from costly burden in practical setting. Iceberg calving datasets are relatively scarce, which increases the risk of overfitting when training these models on limited time series data. In addition, predictive performance is constrained by the sparse set of input variables, such as solely surface mass balance (SMB), North Atlantic Oscillation (NAO), and Labrador Sea Surface Temperature (LSST)~\cite{bigg2014century}. Some performance bias may arise from unobserved key variables that are missing from the modeling process~\cite{zhao2016inferring}. Therefore, this poses a critical question: how to enhance the predictive performance for iceberg discharge without requiring more data samples or informative input variables?

To address the gap, we propose a Hybrid NARX--LLM framework for iceberg discharge forecasting, where a NARX model yields baseline predictions and a large language model (LLM) performs residual correction. Our core premise is that LLMs possess latent reasoning capabilities to integrate glaciological physics into residual correction, thereby enabling data-efficient predictive improvement. However, much of this physical knowledge is qualitative and unstructured, making it difficult to convert into structured prompts that are easily understood by LLMs. 

To this end, we introduce a Physics-Informed Prompt (PIP) method that encodes qualitative glaciological knowledge as template-based natural language instructions to guide the correction process. By feeding these encoded PIPs into the LLM, the model leverages zero-shot in-context reasoning to generate interpretable, physically-grounded reasoning paths, from which corrective residuals are subsequently derived. This approach provides an explicit logical trace for each prediction, significantly improving the physical interpretability of residual correction. Coupled with this interpretive advantage, the LLM enables trend-aware residual correction to assist iceberg discharge modeling. While the naive NARX model is restricted to the short-term temporal mapping of numerical variables within its time steps, the LLM is comparatively more sensitive to long-term physical trends, allowing it to provide specific output corrections that align short-term predictions with the underlying physical trajectory.

The main contributions are summarized as follows:

\begin{itemize}
	\item \textbf{A hybrid NARX--LLM framework:} We propose a novel framework for iceberg discharge forecasting that leverages the numerical mapping strength of NARX and the trend-aware reasoning of LLM to refine baseline predictions via residual correction. This approach suggests a data-efficient paradigm for climate modeling refinement without the need for additional training data or more informative input features.
	\item \textbf{A Physics-Informed Prompt (PIP) method:} We devise a template-based prompting module to encode qualitative glaciological knowledge into structured natural language instructions, enabling the integration of domain physics and baseline behaviours.
	\item \textbf{Enhanced interpretability and robustness:} We demonstrate that our approach provides an explicit logical trace for each prediction via physically-grounded reasoning paths, significantly improving inference interpretability and robustness under non-stationary and partially observed dynamics.
\end{itemize}

\section{Related Works}
This section reviews recent developments in iceberg discharge modeling, residual correction techniques, and the emerging field of LLM-assisted climate forecasting. 

\subsection{Iceberg Discharge Modeling}\label{AA}
Modeling the mass balance of the Greenland Ice Sheet (GrIS) requires an accurate representation of calving discharge, which has been identified as a primary driver of mass loss over the last century~\cite{bigg2014century}. While this discharge is governed by a nonlinear combination of surface mass balance (SMB) and ocean-atmosphere forcing, its predictability is heavily constrained by local glacier dynamics. For instance, as noted by~\cite{benn2017glacier}, the delivery of ice to the ocean is strongly modulated by fjord bathymetry, specifically through the stabilizing effect of pinning points or the instability of over-deepened basins. Current physical ice-sheet models often rely on simplified 'calving laws' that fail to capture these complex interactions, necessitating the development of more robust models to reduce uncertainty in future projections. 

Despite the progress in physics-based modeling~\cite{nick2013future,lea2014terminus,ultee2017plastic}, several studies have pivoted toward data-driven approaches~\cite{bigg2014century,zhao2016inferring}. However, these empirical models are often prone to overfitting limited time series data and frequently struggle to capture extreme events induced by rare climatic variations or non-stationary trends. To bridge these gaps, a hybrid NARX-LLM framework grounded in glaciological physics presents a promising avenue to mitigate prediction errors and capture complex non-linearities by leveraging the contextual reasoning of Large Language Models (LLMs).

\subsection{Residual Correction}
Residual correction methods refine model outputs by systematically modeling previous prediction errors~\cite{friedman2001greedy}. Whether employed as an iterative optimization strategy in boosting frameworks~\cite{chen2016xgboost} or as a dynamic calibration tool in time-series forecasting~\cite{kim2022residual}, this approach enables models to compensate for initial deficiencies. By focusing on the unexplained variance, residual correction significantly enhances the precision and reliability of complex predictive tasks~\cite{lim2021time}. 

While Large Language Models (LLMs) struggle with the direct numerical estimation of iceberg discharge, owing to data scarcity and the inherent instability of pure statistical mappings under non-stationary dynamics~\cite{tang2025time,wang2025novel}, their general temporal pattern extrapolation capabilities~\cite{gruver2023large} offer significant potential as a residual correction mechanism by refining complex non-linearities and long-term trajectories that traditional base models tend to overlook. To harness this potential within glaciological boundaries, we incorporate multivariate physical temporal features, such as Labrador Sea surface temperature (LSST), surface mass balance (SMB), and North Atlantic Oscillation (NAO), into the LLM-based residual correction pipeline.

\subsection{LLM-assisted Climate Forecasting}
Large Language Models (LLMs)~\cite{chen2025integrating} are evolving beyond text-based chatbots to actively assist in climate modeling. Recent work~\cite{cao2024llm} suggests LLMs can bridge the gap between raw data and policy-making by automating scientific workflows, while frameworks like ClimaQA~\cite{manivannan2025climaqa} have emerged to evaluate their scientific reliability. Furthermore, agentic systems such as Zephyrus~\cite{varambally2025zephyrus} demonstrate the efficacy of iterative reflection for complex meteorological tasks. 

Extending these advancements to glaciology, our study focuses on the predictive task of iceberg discharge, leveraging LLMs to refine estimations at a bottleneck where conventional temporal modeling is constrained by dramatic non-stationarity. To the best of our knowledge, this work represents the first attempt to integrate LLM reasoning into iceberg discharge forecasting.

\section{Methodology}
\begin{figure*}[t]
	\centering
	\includegraphics[width=1.06\textwidth]{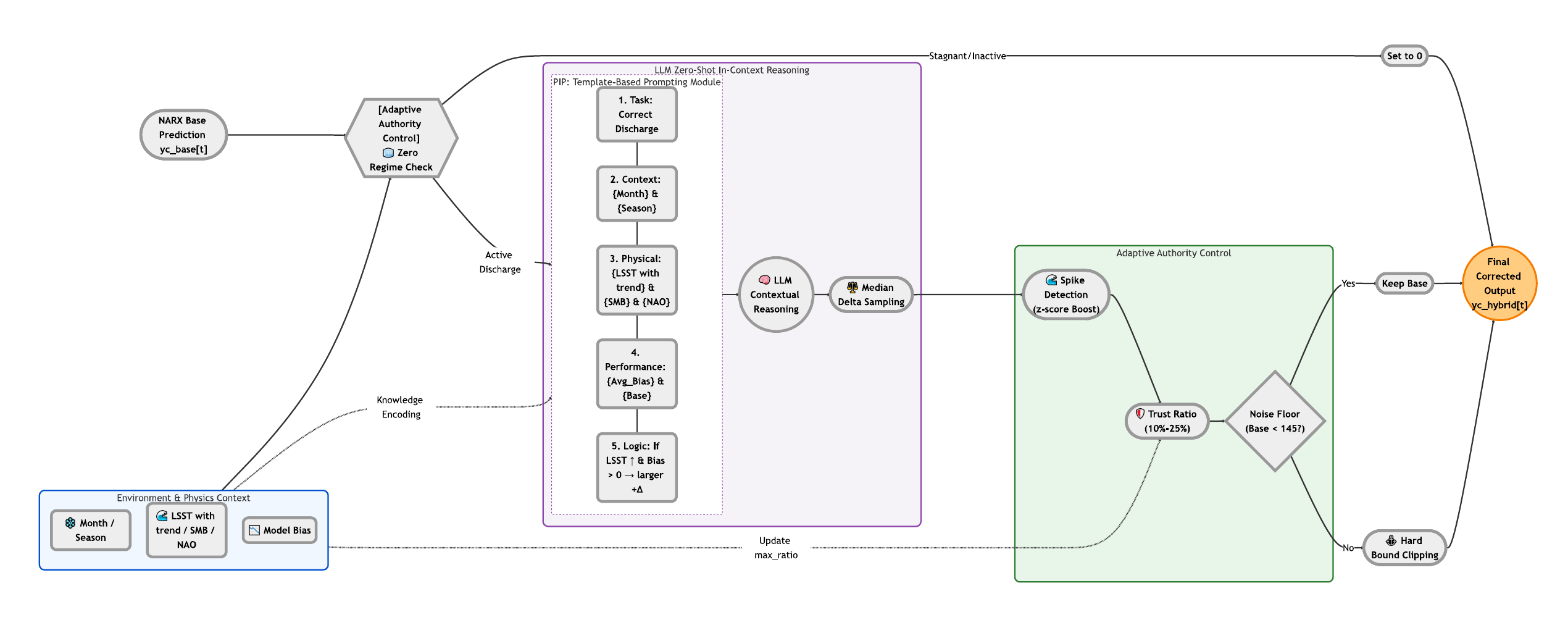}
	\caption{\textbf{The proposed hybrid NARX-LLM framework for Greenland iceberg discharge prediction (Zoom in for detailed view):} The architecture illustrates the integration of a NARX baseline with LLM-based contextual reasoning, featuring: (1) physical sensing of climate states, trend and bias, (2) qualitative glaciological knowledge encoding via PIP, and (3) adaptive authority control consisting of zero regime check, spike detection, dynamic trust ratios and noise floor filtering. }
	\label{fig:narx-llm}
\end{figure*}

This section details the proposed hybrid NARX-LLM framework (Fig.~\ref{fig:narx-llm}), which integrates NARX modeling, the Physics-Informed Prompt (PIP) method, LLM Zero-Shot In-Context Reasoning (ICR), and a suite of Adaptive Authority Control (AAC) modules encompassing the Zero Regime Check, Spike Detection, and robust bounding mechanisms.

\subsection{NARX Modeling}
\label{subsec:narx_modeling}
The Nonlinear Autoregressive Exogenous (NARX) model is employed as the foundational numerical backbone of our framework to capture the standard non-linear temporal dependencies. Given the historical target outputs $I48N(t-1), \dots, I48N(t-d)$ and exogenous physical inputs $\mu(t), \dots, \mu(t-d), re(t), \dots, re(t-d)$, the base NARX prediction $\hat{y}_{\,\text{base}, t}$ is formulated as:
\begin{equation}
	\begin{split}
		\hat{y}_{\,\text{base}, t} = f\big( & I48N(t-1), \dots, I48N(t-d), \\
		& \mu(t), \dots, \mu(t-d), \\
		& re(t), \dots, re(t-d)\big)
	\end{split}
\end{equation}
where $f(\cdot)$ denotes the mapping function, $t$ represents the discrete time step, and $d$ signifies the lag interval. Additionally, $\mu(t)$ and $re(t)$ represent the fixed-effect and random-effect vectors, respectively, constructed from external physical multivariates (SMB, NAO, and LSST) as follows:
\begin{equation}
	\mu(t) = [SMB_\mu(t), NAO_\mu(t), LSST_\mu(t)]
\end{equation}
\begin{equation}
	re(t) = [SMB_{re}(t), NAO_{re}(t), LSST_{re}(t)]
\end{equation}
where $SMB(t)$, $NAO(t)$, and $LSST(t)$ represent the practical observations of each input variable at time index $t$. Their corresponding annual means (fixed effects) are denoted as $SMB_{\mu}(t)$, $NAO_{\mu}(t)$, and $LSST_{\mu}(t)$ in the year corresponding to $t$, while their random effects are expressed as deviations from the mean:
\begin{equation}
	\begin{aligned}
		SMB_{re}(t) &= SMB(t) - SMB_\mu(t) \\
		NAO_{re}(t) &= NAO(t) - NAO_\mu(t) \\
		LSST_{re}(t) &= LSST(t) - LSST_\mu(t)
	\end{aligned}
\end{equation}
Through this formulation, we essentially perform climatological anomaly decomposition on the exogenous features \cite{fettweis2017reconstructions, trenberth1984some}. Guided by the proven efficacy of this method in isolating stochastic physical signals from background states \cite{rb1990stl, kashinath2021physics}, each feature is explicitly decomposed into its annual mean ($\mu$) and residual anomaly ($re$). This strategy effectively separates yearly fixed effects, allowing the model to better capture the yearly random effects that drive critical ice sheet variability.

While NARX modeling excels at capturing steady-state numerical regressions under stable conditions, it exhibits fatal limitations in highly dynamic environments. Specifically, it lacks semantic context awareness of physical information and fails to generalize under severe sudden variations or environmental noise, leading to rigid drift errors. This performance bottleneck directly motivates the integration of our Physics-Informed Prompt method and LLM Zero-Shot In-Context Reasoning module to adaptively refine the raw numerical predictions.

\subsection{Physics-Informed Prompt Method and LLM Zero-Shot In-Context Reasoning}
\label{subsec:pip_icl}

\textbf{Physics-Informed Prompt Method.} Guided by the discovery of the Reasoning Manifold whereby prompt design is key to converging LLM reasoning into an unseen, domain-specific, and minimal low-dimensional structure~\cite{ma2026reasoning}, we propose a specialized Physics-Informed Prompt (PIP) method to systematically bridge the gap between traditional numerical modeling and generative AI models. Rather than treating the Large Language Model (LLM) as a generic statistical regressor, the semantic prompt space is explicitly engineered to encapsulate multi-sphere climate dynamics alongside the baseline model's historical performance metrics. This method encodes raw continuous environmental observations and unstructured glaciological knowledge into a highly structured, domain-specific prompt template, allowing the attention mechanisms within the LLM to effectively evaluate physical boundary conditions and cross-sphere feedback loops in a latent and compressed search space.

The structural breakdown of the PIP construction, along with the corresponding dynamic variables and their specific glaciological  and algorithmic rationales, is comprehensively detailed in Table~\ref{tab:prompt_mapping}. By segmenting the prompt into distinct semantic blocks (\textit{Temporal Boundary}, \textit{Physical State}, \textit{Model Performance}, and \textit{Domain-specific Prior}), we ground the LLM's predictive search space within acceptable physical constraints. Formally, at inference time step $t$, the semantic blocks illustrated in Table~\ref{tab:prompt_mapping} are mathematically formalized to govern the language model's predictive search space.

\begin{table*}[h!] 
	\centering
	\small 
	\caption{Mapping of the Physics-Informed Prompt (PIP) Construction to Dynamic Climate Variables and Glaciological /Algorithmic Rationales.}
	\label{tab:prompt_mapping}
	\setlength{\tabcolsep}{6pt} 
	
	\noindent 
	\hspace*{-0.6cm}%
	\begin{tabularx}{1.15\textwidth}{>{\raggedright\arraybackslash\ttfamily}p{4.7cm} >{\raggedright\arraybackslash\ttfamily}p{2.3cm} X} 
		\toprule
		\textbf{\textrm{PIP Construction}} & \textbf{\textrm{Dynamic Variables}} & \textbf{\textrm{Glaciological Rationale \& \newline Machine Learning Intent}} \\
		\midrule
		"Task: Correct Greenland Iceberg Discharge NARX model prediction." & None & \textbf{Task Grounding:} Establishes the residue-modeling objective, defining the target boundary (Greenland iceberg discharge) and the baseline architecture (NARX). \\
		\addlinespace
		"Current Context: Month \{$m_t$\}, Season: \{\text{Summer}/\text{Winter}\}" & $m_t$, $\mathbb{I}_{\text{summer}}(m_t)$ & \textbf{Temporal Boundary:} Captures the strong seasonal cycle of calving (e.g., summer meltwater lubrication vs. winter sea-ice buttressing). \\
		\addlinespace
		"Physical State: LSST=\{$LSST_t$\}, (Trend=\{$\Delta LSST_t$\}), SMB=\{$SMB_t$\}, NAO=\{$NAO_t$\}" & $LSST_t$, $\Delta LSST_t$, $SMB_t$, $NAO_t$ & \textbf{Key Physical Climate Factors:} \newline 
		$\bullet$ \textit{Labrador Sea 
			Surface Temperature (LSST) \& Trend}: Ocean thermal energy driving marine undercutting and calving. \newline 
		$\bullet$ \textit{Surface Mass Balance (SMB)}: Surface mass loss accelerating glacial sliding. \newline 
		$\bullet$ \textit{North Atlantic Oscillation (NAO)}: Large-scale atmospheric circulation driver. \\
		\addlinespace
		"Model Performance: Average bias in last \{$W$\} months is \{$\bar{e}_t$\}" & $W$, $\bar{e}_t$ & \textbf{Error Adaptation Feedback:} Provides the model's recent moving-average systematic drift to help the LLM calibrate unmodeled physical trends. \\
		\addlinespace
		"Base Prediction: \{$\hat{y}_{\,\text{base}, t}$\}" & $\hat{y}_{\,\text{base}, t}$ & \textbf{Mathematical Anchor:} Supplies the raw, uncorrected output from the NARX model as the baseline benchmark for the delta shift. \\
		\addlinespace
		"Domain-specific Prior: If LSST is rising and bias is positive, a larger positive correction is likely needed." & None & \textbf{Domain Heuristics:} Injects an expert prior rule, forcing the LLM's self-attention mechanisms to correlate ocean warming with NARX model underestimation. \\
		\addlinespace
		"Predict numeric correction delta. Respond ONLY with the number." & None & \textbf{Operational Constraint:} Eliminates conversational stochasticity to guarantee robust downstream float parsing. \\
		\addlinespace
		"Correction delta:" & None & \textbf{Generation Trigger:} Serves as a clear completion token to immediately initiate numerical token prediction. \\
		\bottomrule
	\end{tabularx}
\end{table*}

The \textit{Temporal Boundary} block captures the strong seasonal cycle of calving by mapping the current month $m_t$ to a binary indicator $\mathbb{I}_{\text{summer}}(m_t)$, defined as:
\begin{equation}
	\label{eq:summer_indicator}
	\mathbb{I}_{\text{summer}}(m_t) = 
	\begin{cases} 
		1, & \text{if } m_t \in \{6, 7, 8, 9\}, \\
		0, & \text{otherwise},
	\end{cases}
\end{equation}
where $\mathbb{I}_{\text{summer}}(m_t) = 1$ explicitly maps to the prompt text \texttt{"Summer"}, denoting the period of intense summer meltwater lubrication, whereas $\mathbb{I}_{\text{summer}}(m_t) = 0$ maps to \texttt{"Winter"}, characterizing winter sea-ice buttressing behaviors.

The \textit{Physical State} block encapsulates the concurrent multi-sphere climate variables, comprising the Labrador Sea Surface Temperature ($LSST_t$), Surface Mass Balance ($SMB_t$), and the North Atlantic Oscillation index ($NAO_t$). Within this context boundary, the transient ocean warming velocity (i.e., LSST trend) is computed via a first-order temporal backward difference:
\begin{equation}
	\Delta LSST_t = LSST_t - LSST_{t-1}
\end{equation}

The \textit{Model Performance} block quantifies the systemic drift of the uncorrected baseline prediction $\hat{y}_{\,\text{base}, t}$. This historical error pattern is captured by calculating the average rolling bias $\bar{e}_t$ over a temporal lookback window $W$ against the true observed discharge $y_p$:
\begin{equation}
	\label{eq:average_rolling_bias}
	\bar{e}_t = \frac{1}{W} \sum_{p=t-W}^{t-1} \left( y_p - \hat{y}_{\,\text{base}, p} \right)
\end{equation}

The explicit addition of the domain-specific prior block serves as an abstract conditioning anchor. It guides the LLM's self-attention layers to correlate the thermodynamic trend (\textit{Physical State}) with the statistical behavior of the baseline model (\textit{Model Performance}). To achieve this, the \textit{Domain-specific Prior} block maps these physical and statistical inputs into an explicit expert rule within the prompt template. This domain heuristic guides the language model's reasoning across the structured textual context, forcing it to directly associate positive ocean warming trends ($\Delta LSST_t > 0$) with systematic baseline underestimation ($\bar{e}_t > 0$), mirroring causal dependencies heavily documented in glaciological observations \cite{straneo2013north, moon201221st}.

Consequently, our PIP method acts as a structural knowledge bridge, enabling the text-based architecture to effectively interpret complex environmental anomalies and seamlessly route the coupled physical context to the subsequent In-Context Reasoning pipeline.

\textbf{LLM Zero-Shot In-Context Reasoning.} Within the proposed hybrid NARX-LLM framework, the large language model functions as a trend-aware reasoning agent that generates a residual correction scalar $\delta$ to refine the baseline prediction. Given an input context vector constructed via the PIP method, the robust numerical decoding pipeline operates through a unified formulation. By omitting time index \(t\) for brevity, let $\hat{y}_{\,\text{base}} \in \mathbb{R}$ represent the uncorrected NARX prediction, $\bar{e}$ denote the rolling historical average bias computed over a temporal lookback window $W$, $m \in \{1, 2, \dots, 12\}$ represent the seasonal month index, $\mathbf{s} \in \mathbb{R}^4$ denote the concurrent physical climate vector, and $\mathcal{R}$ denote the expert heuristic rules. The PIP operator $\mathcal{P}(\cdot)$ aggregates these heterogeneous features into a unified textual prompt vector:
\begin{equation}
	\mathbf{x}_{\text{PIP}} = \mathcal{P}\left(\hat{y}_{\,\text{base}}, \mathbf{s}, \bar{e}, W, m, \mathcal{R}\right)
\end{equation}

Crucially, we intentionally formulate this pipeline as a zero-shot inference scheme rather than relying on cross-instance few-shot exemplars. This design systematically mitigates the risk of compounding numerical noise from mismatched historical cases, drastically reduces token consumption for continuous forecasting deployment, and isolates the LLM's capacity to generalize solely under the governing physical heuristics $\mathcal{R}$.

To circumvent the model's epistemic uncertainty under autoregressive generation and introduce structural error-tolerance against single-path generation freezing, the language model $\mathcal{M}_{\text{LLM}}$ is conditioned on the context string $\mathbf{x}_{\text{PIP}}$ to concurrently sample a set of $N=3$ independent response trajectories $\mathcal{O} = \{O_i\}_{i=1}^{N}$ according to the conditional generation space:
\begin{equation}
	O_i \sim \mathcal{M}_{\text{LLM}}\left( \cdot \mid \mathbf{x}_{\text{PIP}}; \, \tau = 0.1 \right)
\end{equation}
where decoding hyperparameters are strictly bounded to a low temperature ($\tau = 0.1$) with random sampling enabled (\texttt{do\_sample=True}) and a restricted token cap ($L_{\text{max}} = 15$). This localized exploration parameterization guarantees candidate diversity while preventing unbounded semantic drift.

Following sequence generation, a deterministic parsing operator $\mathcal{E}(\cdot)$ maps each text trajectory $O_i$ to its corresponding numerical target attribute, filtering out descriptive text fragments to isolate the targeted scalar $c_i = \mathcal{E}(O_i)$. The valid real-valued extractions are consolidated into a filtered candidate set:
\begin{equation}
	\mathcal{C} = \{c_i \mid c_i = \mathcal{E}(O_i), \, c_i \in \mathbb{R}\}
\end{equation}
where the sequence cardinality satisfies $0 \le |\mathcal{C}| \le N$. To protect downstream tasks from systemic formatting anomalies or heavy-tailed decoding failures, an instance-level drop mechanism (\texttt{continue}) is executed if $\mathcal{C} = \emptyset$. For non-empty candidate sets, the final predictive correction $\delta$ is established using a non-linear median reduction function acting as the aggregation layer:
\begin{equation}
	\delta = \text{median}\left( \mathcal{C} \right)
\end{equation}
By prioritizing the non-linear median operator over empirical averages, the aggregation layer provides robust breakdown-point protection against heavy-tailed linguistic anomalies or orders-of-magnitude hallucinations, inherently neutralizing unexpected numerical fluctuations and preserving structural fidelity across continuous timeline intervals.

\subsection{Adaptive Authority Control Modules}
\label{subsec:aac_modules}

To ensure numerical stability and insulate the framework pipeline from generative variance, we implement the Adaptive Authority Control (AAC) modules. As illustrated in Fig.~\ref{fig:narx-llm}, AAC modules consist of Zero Regime Check (ZRC), Spike Detection (SD), Dynamic Trust Region (DTR), and Noise-Floor Saturation Bounding (NFSB).

\noindent\textbf{Zero Regime Check:} 
To eliminate redundant LLM calls and suppress background noise, at each temporal step $t$, the historical window vector $\mathbf{y}_{\text{true}, t-W:t}$ is sliced from the ground-truth observations $y_{\,\text{true}}$ based on a lookback size $W$:
\begin{equation}
	\mathbf{y}_{\text{true}, t-W:t} = \left[ y_{\,\text{true}}[k] \right]_{k=\max(0, \, t-W)}^{t-1}
\end{equation}
By evaluating the first and second moments over this windowed vector, the Zero Regime Check $\mathbb{I}_{\text{ZRC}}\left(\cdot \right)$ is defined as follows:
\begin{equation}
	\mathbb{I}_{\text{ZRC}}\left( \mathbf{y}_{\text{true}, t-W:t} \right) = \begin{cases} 
		1, & \text{if } \text{Mean}(\mathbf{y}_{\text{true}, t-W:t}) < \tau_{\mu} \\
		& \quad \wedge \text{ Std}(\mathbf{y}_{\text{true}, t-W:t}) < \tau_{\sigma} \\
		0, & \text{otherwise}
	\end{cases}
\end{equation}
where the thresholds $\tau_{\mu}$ and $\tau_{\sigma}$ are empirically set to $0.15$ and $0.03$ respectively, to isolate baseline measurement noise.
If $\mathbb{I}_{\text{ZRC}}\left(\cdot \right) = 1$, the current state is classified as \textit{Stagnant/Inactive}. Under this regime, the prediction process skips the generative layer entirely via an instance-level drop mechanism (\texttt{continue}), and the final output is directly assigned as $y_{\,\text{hybrid},t} = 0$. Otherwise, the system triggers the \textit{Active Discharge} path and authorizes the downstream LLM pipeline.

Regarding the \textit{Active Discharge} path, the raw response trajectories are pooled into an initial correction candidate $\delta_t = \text{median}\left( \mathcal{C}_t \right)$ via the aggregation layer. The post-inference AAC applies three sequential modules to regularize this correction value:

\textbf{Spike Detection ($Z$-score Boost):} To prevent baseline model underestimation during sudden shifts, a rolling $Z$-score is computed over a historical window of length $W_s$ ($W_s=7$):
\begin{equation}
	z_t = \frac{\hat{y}_{\,\text{base}, t} - \mu_t}{\max(\sigma_t, \, \sigma_{\min}) + \epsilon}
\end{equation}
where $\mu_t = \operatorname{Mean}(\mathbf{y}_{\text{true}, t-W_s:t})$ and $\sigma_t = \operatorname{Std}(\mathbf{y}_{\text{true}, t-W_s:t})$. Once a sharp peak is encountered ($|z_t| > \tau_z$), the LLM-inferred residual correction scalar $\delta_t$ is dynamically amplified to overcome generative reasoning and tracking lag:
\begin{equation}
	\delta_t \leftarrow \begin{cases} 
		\delta_{t} \cdot \left[1 + \gamma \cdot \min(|z_t| - \tau_z, \, \kappa)\right], & \text{if } |z_t| > \tau_z \\
		\delta_{t}, & \text{otherwise}
	\end{cases}
\end{equation}
where the hyperparameters $\sigma_{\min}$, $\epsilon$, $\tau_z$, $\gamma$, and $\kappa$ are empirically set to $0.1$, $10^{-6}$, $2.5$, $0.5$, and $2.0$, respectively.

\textbf{Dynamic Trust Region:} To insulate the hybrid trajectory from unbounded generative variance and enhance baseline tracking stability, a dynamic saturation envelope is enforced on the residual correction. We define an extreme glaciological indicator $\mathbb{I}_{\text{ext}, t} \in \{0, 1\}$ based on climate observations:
\begin{equation}
	\mathbb{I}_{\text{ext}, t} = \begin{cases} 
		1, & \text{if } \text{LSST}_t > \mathcal{P}_{80}(\{\text{LSST}_k\}_{k=1}^N) \\
		& \text{or } \text{SMB}_t < \mathcal{P}_{20}(\{\text{SMB}_k\}_{k=1}^N) \\
		0, & \text{otherwise}
	\end{cases}
\end{equation}
where $\text{LSST}_t$ and $\text{SMB}_t$ represent the Labrador Sea Surface Temperature and Surface Mass Balance at time $t$, respectively. The sequences $\{\text{LSST}_k\}_{k=1}^N$ and $\{\text{SMB}_k\}_{k=1}^N$ denote the full historical observation ensembles across all $N$ time steps in the training partition of the dataset, and $\mathcal{P}_k(\cdot)$ outputs the $k$-th percentile of the targeted ensemble. The dynamic ceiling $\alpha_t$ is modulated by environmental conditions and directional alignment with average rolling bias $\bar{e}_t$ defined in Eq.~\ref{eq:average_rolling_bias}:
\begin{equation}
	\label{eq:dynamic_ceiling}
	\begin{split}
		\alpha_t = \, & \max\left(\alpha_0, \, \Delta \alpha_s \mathbb{I}_{\text{summer}}(m_t), \, \Delta \alpha_e \mathbb{I}_{\text{ext}, t}\right) \\
		& + \Delta \alpha_b \cdot \mathbb{I}\left(\operatorname{sgn}(\delta_t) = \operatorname{sgn}(\bar{e}_t)\right)
	\end{split}
\end{equation}
where $\alpha_0 = 0.10$, $\Delta \alpha_s = 0.20$, $\Delta \alpha_e = 0.25$, $\Delta \alpha_b = 0.10$ and $\mathbb{I}_{\text{summer}}(m_t)$ is defined in Eq.~\ref{eq:summer_indicator}. This adaptive ceiling restricts the generative correction $\delta_t$ via a standard clipping operator:
\begin{equation}
	\delta^*_t = \operatorname{clip}\left(\delta_t, \, -\alpha_t |\hat{y}_{\,\text{base}, t}|, \, \alpha_t |\hat{y}_{\,\text{base}, t}|\right).
\end{equation}

\textbf{Noise-Floor Saturation Bounding:} The final integrated prediction $\hat{y}_{\,\text{hybrid}, t}$ is conditionally activated based on a baseline threshold $\theta_{\text{floor}} = 145$ and a core correction tolerance $\tau_{\delta} = 0.1$. If $\hat{y}_{\,\text{base}, t} < \theta_{\text{floor}}$ or $|\delta^*_t| \le \tau_{\delta}$, the framework disables generative corrections entirely by setting $\hat{y}_{\,\text{hybrid}, t} = \hat{y}_{\,\text{base}, t}$. Otherwise, to capture impending massive calving events, the constrained correction $\delta^*_t$ is conditionally accelerated to an amplified $\tilde{\delta}_t$:
\begin{equation}
	\label{eq:amplified_delta}
	\tilde{\delta}_t = \begin{cases}
		2\delta^*_t, & \text{if } \hat{y}_{\,\text{base}, t} + \delta^*_t > \\
		& \quad \mathcal{P}_{99}(\{\hat{y}_{\,\text{base}, k}\}_{k=t-W_b}^{t-1}) \\
		\delta^*_t, & \text{otherwise}
	\end{cases}
\end{equation}
where $\mathcal{P}_{99}(\cdot)$ denotes the 99th percentile of the sequence of baseline predictions within a historical window of $W_b=10$. The final trajectory is safely aggregated via a standard clipping operator:
\begin{equation}
	\label{eq:final_hybrid}
	\hat{y}_{\,\text{hybrid}, t} = \operatorname{clip}\left(\hat{y}_{\,\text{base}, t} + \tilde{\delta}_t, \; 0, \; \lambda \cdot y_{\text{train\_max}}\right)
\end{equation}
where $\lambda = 1.2$, and $y_{\text{train\_max}}$ denotes the maximum historical discharge volume recorded in the training ensemble.

\section{Experiment and Discussion}

This section outlines the dataset specifications, implementation setup, experimental results, and performance analysis, followed by a series of ablation studies.

\subsection{Dataset and Implementation Setup}
\noindent\textbf{Dataset Description.} We utilize a historical climate-glaciological dataset spanning from 1901 to 2018, which is an extended version obtained from the authors of~\cite{bigg2014century}. The target output variable is $I48N$, which represents the monthly counts of icebergs crossing $48^\circ\text{N}$ in the West Atlantic, sourced from the U.S. Coast Guard's International Ice Patrol. Serving as a robust first-order proxy for iceberg discharge from South and West Greenland, this output is modeled against three key environmental forcing inputs: Surface Mass Balance (SMB) representing glaciological drivers, the North Atlantic Oscillation (NAO) tracking atmospheric dynamics, and the Labrador Sea Surface Temperature (LSST) capturing oceanic variations. All variables are aligned at a monthly temporal resolution to capture the time-varying correlations across the 118-year observation range. 

To capture temporal dependencies, the dataset is structured into a Non-linear Autoregressive with Exogenous Inputs (NARX) format with a time delay of $d=12$ months. Finally, the sequential data is chronologically split into a training set and a testing set using an 80:20 ratio to prevent data leakage and ensure the model does not look ahead into future values.

\noindent\textbf{Implementation Details.} The hybrid NARX-LLM framework is implemented in PyTorch, deploying the \texttt{Qwen2.5-1.5B-Instruct} model~\cite{yang2024qwen2} locally on a single NVIDIA GPU using half-precision (FP16) to maximize throughput and memory efficiency. This 1.5B-parameter backbone delivers a lightweight, edge-deployable solution with a minimal VRAM footprint, ensuring the framework remains computationally practical for long-term earth science forecasting without demanding high-end supercomputing infrastructure. The temporal lookback window $W$ in Table~\ref{tab:prompt_mapping} is set to 3. Following the standard formulation of the NARX model~\cite{lin1996learning}, the baseline constructs an input vector of 12 historical lags from both exogenous and target variables. Its architecture comprises a single hidden layer (10 units, Tanh activation) followed by a 0.2 dropout layer, trained over 2,000 epochs to minimize Mean Squared Error (MSE) via the Adam optimizer ($lr = 0.005$, $\text{weight decay} = 1 \times 10^{-4}$). All evaluations are conducted across 5 independent runs with different random seeds, and the mean performance is reported to mitigate the impact of stochastic volatility and ensure reproducibility.

To rigorously evaluate the predictive performance, we employ four standard metrics: Root Mean Squared Error (RMSE), Mean Absolute Error (MAE), the Coefficient of Determination ($R^2$), and the Explained Variance Score ($\text{EVS}$). While RMSE and MAE measure absolute deviations, $R^2$ and $\text{EVS}$ evaluate the proportion of variance captured by the models. These metrics are mathematically formulated as follows:

\begin{equation}
	\text{RMSE} = \sqrt{\frac{1}{N} \sum_{t=1}^{N} (y_{\,\text{true},t} - \hat{y}_{\mathcal{M},t})^2}
\end{equation}
\begin{equation}
	\text{MAE} = \frac{1}{N} \sum_{t=1}^{N} |y_{\,\text{true},t} - \hat{y}_{\mathcal{M},t}|
\end{equation}
\begin{equation}
	R^2 = 1 - \frac{\sum_{t=1}^{N} (y_{\,\text{true},t} - \hat{y}_{\mathcal{M},t})^2}{\sum_{t=1}^{N} (y_{\,\text{true},t} - \bar{y}_{\,\text{true}})^2}
\end{equation}
\begin{equation}
	\text{EVS} = 1 - \frac{\text{Var}(y_{\,\text{true},t} - \hat{y}_{\mathcal{M},t})}{\text{Var}(y_{\,\text{true},t})}
\end{equation}
where $\mathcal{M} \in \{\text{base}, \text{hybrid}\}$ denotes the model type, $N$ is the total number of test instances, and $y_{\,\text{true},t}$ represents the observed monthly iceberg count at time step $t$. The term $\hat{y}_{\mathcal{M},t}$ is the prediction yielded by model $\mathcal{M}$, $\bar{y}_{\,\text{true}}$ represents the empirical mean of the ground truth, and $\text{Var}(\cdot)$ denotes the sample variance operator. Higher $R^2$ and $\text{EVS}$ values, along with lower $\text{RMSE}$ and $\text{MAE}$ values, indicate superior predictive accuracy.

\subsection{Experimental Results and Performance Analysis}

\begin{table}[htbp]
	\caption{Quantitative performance comparison between Base NARX and Hybrid NARX-LLM models. All results are reported as the mean across 5 independent randomized runs. The best performing values are highlighted in \textbf{bold}, with the percentage reduction of MAE and standard deviation ($\pm$) of $R^2$ explicitly annotated.}
	\label{tab:performance_summary}
	\centering
	\begin{tabular}{lll}
		\toprule
		\textbf{Metric} & \multicolumn{1}{c}{\textbf{NARX}} & \multicolumn{1}{c}{\textbf{Hybrid NARX-LLM}} \\ \midrule
		RMSE $\downarrow$ & 92.2183 & \textbf{91.6311} \\
		MAE $\downarrow$  & 44.1187 & \textbf{37.5294}\rlap{\quad (\textbf{-14.94\%})} \\
		EVS $\uparrow$  & 0.4485  & \textbf{0.4532} \\
		$R^2$ $\uparrow$ & 0.4461  & \textbf{0.4532}\rlap{\quad $\pm$ 0.0028} \\ \bottomrule
	\end{tabular}
\end{table}

\begin{figure}[htbp]
	\centering
	\includegraphics[width=0.95\linewidth]{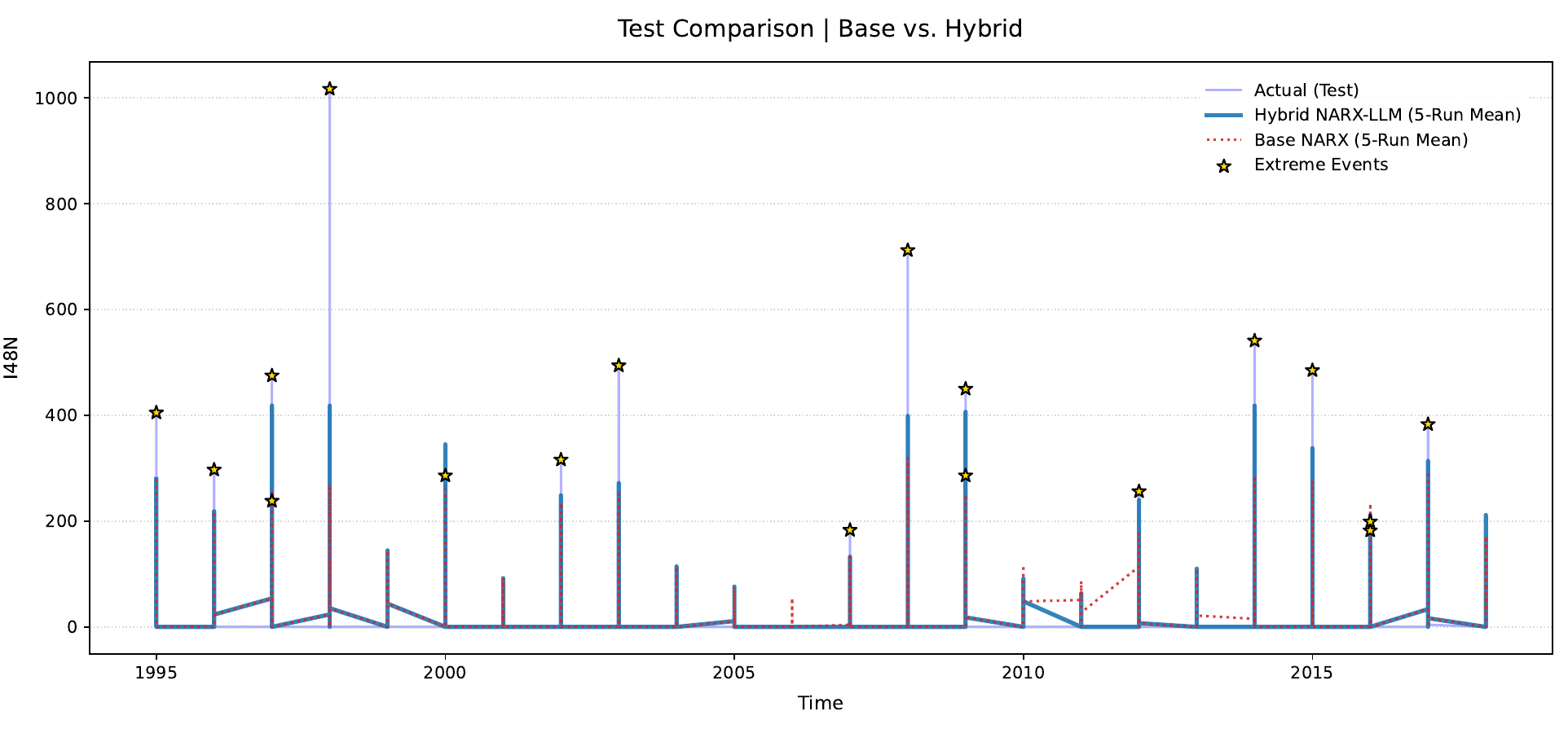} \caption{Qualitative prediction trajectory comparison and extreme event tracking. Both curves represent the 5-run mean performance, where gold stars ($\star$) visually anchor the localized true peaks ($y_{\,\text{true},t} > 150$) to highlight the Hybrid model's capability in enhancing peak-tracking precision. (Zoom in for detailed view)}
	
	\label{fig:test_comparison}
\end{figure}

\noindent\textbf{Quantitative Comparison.} Table~\ref{tab:performance_summary} presents the quantitative comparison between the NARX baseline and hybrid NARX-LLM, evaluated across 5 independent randomized runs. As demonstrated by the empirical outcomes, the hybrid framework structurally outperforms the classical baseline across all evaluated dimensions.

Globally, the hybrid framework effectually pushes both the $R^2$ and EVS metrics to a stable 0.4532, surpassing the baseline scores of 0.4461 and 0.4485, respectively. Most remarkably, the $R^2$ and EVS scores under the hybrid system converge to an identical empirical value of $0.4532$. In statistical modeling, this exact convergence indicates that the average prediction error of the framework is nearly zero ($\bar{y}_{\text{true}} - \bar{\hat{y}}_{\,\text{hybrid}} \approx 0$). This empirical alignment, further verified by the exceptionally low stochastic variance of $\pm 0.0028$ for $R^2$, strongly confirms that the PIP-driven contextual reasoner generates strictly unbiased predictions without introducing systematic variance or variance inflation.

The most prominent optimization is observed in the MAE, where the hybrid NARX-LLM reduces the error from 44.1187 to 37.5294, yielding a statistically significant error reduction of approximately 14.94\%. This distinct contrast between the incremental optimization in RMSE and the substantial reduction in MAE yields a critical methodological insight: while squared metrics are inherently constrained by the intense volatility and extreme peak sparsity of the climate dataset, the PIP-enabled zero-shot reasoning capability enhances the absolute baseline tracking accuracy across the dominant temporal horizons.

\noindent\textbf{Qualitative Trajectory.} To visually validate these statistical improvements, Figure~\ref{fig:test_comparison} illustrates the continuous tracking trajectory across the test timeline, where gold stars ($\star$) denote the extreme events.

The qualitative trajectory analysis reveals two critical advantages of our hybrid system that explain the quantitative error reductions in Table~\ref{tab:performance_summary}. First, the classical NARX baseline, restricted to short-term temporal mapping, systematically suffers from severe peak attenuation, heavily underestimating critical spikes (e.g., the historical maxima in 1998 and 2008) due to extreme data sparsity. Conversely, by tracking continuous trends through the prompt-guided lookback window $W$, the hybrid framework effectively elevates the predictive boundaries toward the empirical ceilings, validating its trend-aware reasoning capability across varying peak magnitudes (including the sharper peaks observed in 1997 and 2009, alongside the milder peak of 2012). Second, during the stable baseline phase between 2010 and 2012, the hybrid framework consistently dampens the false fluctuations predicted by NARX, verifying its dual capability of residual correction in terms of both extreme peak anomalies and spurious fluctuations.

Furthermore, during anomalous periods such as 2016, multiple localized peaks cluster in rapid succession, resulting in a dense concentration of extreme events on the plot. Notably, the hybrid framework adaptively tracks these rapid, continuous physical fluctuations, maintaining its upper-bound trend alignment without falling into chaotic error accumulation. However, minor limitations remain, such as a slight overestimation of the 2000 peak and a persisting gap between the predictions and the 1998 and 2008 historical maxima.

\noindent\textbf{Interpretability of Physically-Grounded Reasoning Paths.} To evaluate interpretability, we qualitatively analyze the explicit logical traces generated via zero-shot in-context reasoning. 
Instead of operating as a black box, our framework forces the LLM to evaluate coupled climatic and glaciological variables prior to predicting corrections, as displayed in Table~\ref{tab:qualitative_traces}.

Cross-examining these cases confirms that the LLM performs physical context modulation rather than blindly replicating historical tracking biases. 
In Case 1 (stable summer, $\text{LSST} = 0.32^\circ\text{C}$, $\text{Trend} = -0.04$), a minor historical deficit is moderated to yield a conservative correction ($\delta = 15.88$), shifting the base prediction ($150.63$) toward the true benchmark ($\text{GT} = 182.00$). 
In Case 2 (accelerated melting, $\text{LSST} = 1.02^\circ\text{C}$, $\text{Trend} = +0.15$), the model detects high-melting conditions and derives a substantial positive residual ($\delta = 77.40$), successfully pushing the hybrid output up to $253.45$ and capturing the massive deficit relative to the benchmark ($\text{GT} = 263.00$). 
Finally, Case 3 illustrates an amplified forcing scenario where temperatures remain mild ($\text{LSST} = 0.33^\circ\text{C}$), but a severe multi-month underestimation persists (Bias: $-28.93$). 
The LLM dynamically balances these coupled dynamics to issue a major correction ($\delta = 79.99$), driving the hybrid output to $239.98$ and bridging over $83\%$ of the base prediction error ($\text{GT} = 256.00$).

This adaptive text-to-numeric modulation verifies that reasoning paths are strictly grounded in physical state constraints. 
By generating explicit logical traces before emitting numerical tokens, the framework guarantees hybrid predictions remain bound to environmental dynamics rather than statistical hallucinations, ensuring the transparency required for critical Earth science deployments.

\begin{table*}[t]
	\centering
	\caption{Representative samples of explicit logical traces and physically-grounded reasoning paths generated via zero-shot in-context reasoning alongside baseline predictions and ground truths. Note: $\delta$ and GT signify delta and ground truth, respectively.}
	\label{tab:qualitative_traces}
	\footnotesize 
	\setlength{\tabcolsep}{3pt} 
	\noindent\hspace*{-1.3cm}%
	\begin{tabular}{lp{4.5cm}p{4.5cm}p{4.5cm}}
		\toprule
		\textbf{Component} & \textbf{Case 1: Stable Ambient Baseline} & \textbf{Case 2: Accelerated Thermal Discharge} & \textbf{Case 3: Amplified Cryospheric Forcing} \\ \midrule
		\textbf{Task Context} & Discharge Correction (Month 6, Summer) & Discharge Correction (Month 7, Summer) & Discharge Correction (Month 8, Summer) \\ \midrule
		\textbf{Physical State} & $\text{LSST}\!=\!0.32^\circ\text{C}$ (Trend: $-0.04$), \newline $\text{SMB}\!=\!42.28$, $\text{NAO}\!=\!0.33$ & $\text{LSST}\!=\!1.02^\circ\text{C}$ (Trend: $+0.15$), \newline $\text{SMB}\!=\!58.47$, $\text{NAO}\!=\!-1.74$ & $\text{LSST}\!=\!0.33^\circ\text{C}$ (Trend: $-0.05$), \newline $\text{SMB}\!=\!30.09$, $\text{NAO}\!=\!-0.83$ \\ \midrule
		\textbf{Base Tracking} & Last 3M Bias: $-6.32$ / Pred: $150.63$ & Last 3M Bias: $-16.30$ / Pred: $176.05$ & Last 3M Bias: $-28.93$ / Pred: $159.99$ \\ \midrule
		\textbf{Reasoning Path} & \multicolumn{3}{p{14.0cm}}{\texttt{If LSST is rising and bias is positive, a larger positive correction is likely needed.}} \\ \midrule
		\textbf{Output Delta} & \texttt{delta: 15.88} & \texttt{delta: 77.40} & \texttt{delta: 79.99} \\
		\textbf{Hybrid Pred.} & \textbf{166.51} & \textbf{253.45} & \textbf{239.98} \\ \midrule
		\textbf{Ground Truth} & \textbf{182.00} & \textbf{263.00} & \textbf{256.00} \\ \bottomrule
	\end{tabular}
\end{table*}

In summary, the collective empirical evidence confirms that by coupling the PIP guidance and trend-aware LLM reasoning with the numerical mapping strength of NARX, our framework achieves robust residual correction while preserving global system stability. Moreover, this integration presents a data-efficient paradigm for climate prediction refinement that obviates the need for additional training data or expanded feature engineering.

\subsection{Ablation Study}

\begin{table}[t]
	\centering
	\caption{Ablation analysis of different Physics-Informed Prompt (PIP) semantic blocks on performance effect. All results are averaged over 5 runs with random seeds. Bold and underlined values indicate the best and second scores per column.}
	\label{tab:ablation_pip}
		\begin{tabular}{lcccc}
			\toprule
			\textbf{Configuration} & \textbf{RMSE} $\downarrow$ & \textbf{MAE} $\downarrow$ & \textbf{EVS} $\uparrow$ & \textbf{$R^2$ $\uparrow$ $\pm$ Std}  \\ 
			\midrule
			Full PIP Semantic Blocks & \underline{91.6311} & 37.5294 & \underline{0.4532} & \underline{0.4532} $\pm$ 0.0028 \\ 
			\quad w/o Temporal Boundary & \textbf{91.3694} & 37.5004 & \textbf{0.4563} & \textbf{0.4563} $\pm$ 0.0042 \\
			\quad w/o Physical State & 97.3483 & 38.1283 & 0.3909 & 0.3826 $\pm$ 0.0197 \\
			\quad w/o Model Performance & 91.9777 & \textbf{36.8876} & 0.4507 & 0.4490 $\pm$ 0.0016 \\
			\quad w/o Domain-specific Prior & 94.8520 & \underline{37.2823} & 0.4206 & 0.4140 $\pm$ 0.0032 \\
			\bottomrule
		\end{tabular}%
\end{table}

\noindent\textbf{Effect of PIP semantic blocks.} As the PIP method partitions its constructed prompt into four semantic blocks (Temporal Boundary, Physical State, Model Performance, and Domain-specific Prior), we investigate the specific impact of each block. 

As shown in Table~\ref{tab:ablation_pip}, removing the Temporal Boundary (\textit{w/o Temporal Boundary}) marginally improves mean metrics (RMSE: 91.3694, $R^2$: 0.4563), but expands the variance from $\pm$0.0028 to $\pm$0.0042, verifying its necessity as a stabilizing regularizer. Conversely, the Physical State block is a critical cornerstone; its removal (\textit{w/o Physical State}) triggers severe structural collapse, plunging $R^2$ to 0.3826 and causing a near-7x variance explosion ($\pm$0.0197) due to LLM hallucinations in the absence of physical constraints. 

Furthermore, removing Model Performance (\textit{w/o Model Performance}) drops MAE to a global best of 36.8876 but degrades RMSE to 91.9777, indicating that it prevents the model from shifting to conservative, mean-centric predictions that ignore heavy-tailed extreme errors. Finally, dropping Domain-specific Prior (\textit{w/o Domain-specific Prior}) diminishes overall accuracy (RMSE: 94.8520, $R^2$: 0.4140), proving that specialized context rules correlating ocean warming with baseline underestimation are indispensable for tracking chaotic phase transitions. Ultimately, the full PIP integrates all blocks to trade marginal performance drops for overall balance with robust global error mitigation and stochastic reliability.

\begin{table}[htbp]
	\centering
	\caption{Ablation analysis of the Adaptive Authority Control (AAC) sub-modules on the hybrid NARX-LLM. Bold and underline indicate the best and second performance within the related variants, where ZRC, SD, DTR, and NFSB denote Zero Regime Check, Spike Detection, Dynamic Trust Region, and Noise-Floor Saturation Bounding, respectively.}
	\label{tab:aac_ablation}
		\begin{tabular}{lccc}
			\toprule
			\textbf{Configuration} & \textbf{RMSE} $\downarrow$ & \textbf{MAE} $\downarrow$ & \textbf{$R^2$ $\uparrow$ $\pm$ Std} \\
			\midrule
			w/o Complete AAC Modules & 92.7189 & 48.1000 & 0.4401 $\pm$ 0.0033 \\
			\quad w/o ZRC & 92.0884 & 44.6975 & 0.4477 $\pm$ 0.0036 \\
			\quad w/o SD & 91.8854 & \underline{37.6196} & 0.4501 $\pm$ 0.0023 \\
			\quad w/o DTR \& NFSB & \textbf{90.7615} & 39.4352 & \textbf{0.4635} $\pm$ 0.0027 \\
			\midrule
			\textbf{Full framework (With All AAC Modules)} & \underline{91.6311} & \textbf{37.5294} & \underline{0.4532} $\pm$ 0.0028 \\
			\bottomrule
		\end{tabular}%
\end{table}

\noindent\textbf{Effect of AAC modules.} To evaluate the contributions of the constituents within the Adaptive Authority Control (AAC), we systematically isolate and investigate the performance impact of omitting each block. As summarized in Table \ref{tab:aac_ablation}, completely bypassing the AAC stabilization mechanism (\textit{w/o Complete AAC Modules}) triggers an immediate and severe performance degradation across all primary evaluation dimensions (MAE inflates to 48.1000, and RMSE climbs to 92.7189), underscoring its indispensable role in preserving structural robustness and numerical stability.

Specifically, the Zero Regime Check (ZRC) and Spike Detection (SD) blocks form the bedrock of baseline precision. Eliminating the ZRC (\textit{w/o ZRC}) leads to a severe degradation in MAE to 44.6975 and a noticeable drop in $R^2$ to 0.4477, demonstrating that failing to properly gate near-zero inactive regimes causes low-level environmental noise to heavily contaminate the residual modeling. Similarly, removing the spike-boosting safeguard (\textit{w/o SD}) yields a clear precision drop, driving the MAE up to 37.6196.

Crucially, the Dynamic Trust Region (DTR) and Noise-Floor Saturation Bounding (NFSB) are treated as a unified boundary constraint assembly (\textit{w/o DTR \& NFSB}) to eliminate localized functional redundancy, as both are jointly engineered to enforce rigid boundary clamping under volatile edge environments. Stripping this joint assembly reveals an instructive performance trade-off: the configuration achieves the optimal empirical RMSE (90.7615) and $R^2$ (0.4635), but at the cost of a degraded mean-error tracking accuracy, which pushes the MAE up to 39.4352. This discrepancy highlights that although the rigid double-protection inherent in the joint assembly imposes a minor regularization penalty under non-extreme data distributions, it is structurally essential for tightening average error bounds. 

Ultimately, the default Full framework consistently secures either the absolute best or second-best metrics across all dimensions. Most notably, it achieves the lowest global average error (MAE: 37.5294), validating that these components successfully synergize to maximize performance reliability for real-world deployments. Furthermore, depending on the characteristics of practical time-series scenarios, selective inclusion or omission of these modules can offer a scalable path toward further performance optimization. 

\begin{figure}[!t]
	\centering
	\includegraphics[width=0.48\textwidth]{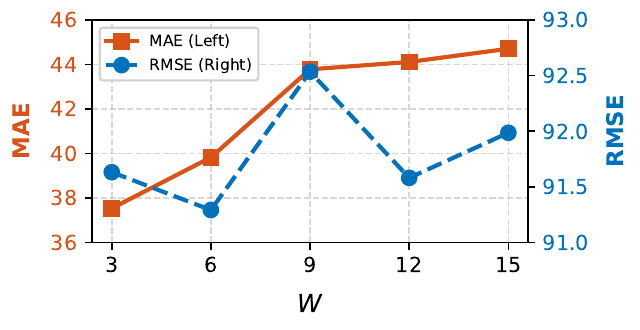} 
	\caption{Evaluation of the hybrid NARX-LLM framework under varying temporal lookback windows $W$. The left and right axes illustrate the trends of MAE and RMSE, respectively.}
	\label{fig:lookback_window}
\end{figure}

\noindent\textbf{Effect of temporal lookback window $W$.} The temporal lookback window $W$ defines the historical context length provided to the framework. We investigate its specific impact on the LLM's trend-aware reasoning capacity by systematically adjusting $W$, thereby assessing how varying historical sequence spans influence the framework's capability to capture complex temporal dynamics. As demonstrated by the empirical results in Fig.~\ref{fig:lookback_window}, the framework achieves its lowest global average error (MAE: 37.5294) at a tighter window size of $W=3$. As $W$ expands further (from 6 up to 15), the MAE consistently degrades, indicating that an excessively prolonged historical context over-extends the temporal horizon and introduces obsolete historical noise that misleads the LLM's trend-reasoning core. Interestingly, a slightly larger window such as $W=6$ yields minor improvements in RMSE (91.2909), highlighting a subtle trade-off where moderate context assists in mitigating peak prediction variances, whereas a concise lookback remains structurally essential for minimizing overall mean errors in highly dynamic iceberg discharge scenarios. Therefore, prioritizing the global suppression of mean errors to maximize real-world deployment reliability, we select $W=3$ as the default configuration for our hybrid framework.

\section{Conclusion}
This paper has presented a hybrid NARX--LLM framework equipped with a Physics-Informed Prompt (PIP) method and Adaptive Authority Control (AAC) modules to address the complex, nonlinear, and poorly observed dynamics of Greenland iceberg discharge. By integrating the structural temporal mapping of NARX with knowledge-driven, LLM-based residual correction, the zero-shot in-context reasoning approach effectively tracks and rectifies rare extreme events and nonstationary trends that typically undermine the traditional predictive method. Consequently, this framework reveals a scalable, interpretable, and robust pathway for bridging data-limited climate forecasting with physics-informed, AI-driven reasoning. Future work will focus on expanding the framework's scalability by incorporating multimodal environmental drivers and evaluating its generalization across other major ice sheets globally.




\section*{Declarations}

\subsection*{Funding}
The authors received no financial support for the research, authorship, or publication of this article.

\subsection*{Competing Interests}
The authors declare that they have no competing financial interests or personal relationships that could have appeared to influence the work reported in this paper.

\subsection*{Data Availability}
The dataset was obtained from a third party and is not publicly available due to proprietary restrictions.

\subsection*{Code Availability}
The source code developed during the current study is available from the corresponding author on reasonable request.

\subsection*{Author Contribution}
\noindent \textbf{Y.G.}: Conceptualization, Methodology, Investigation, Formal analysis, Writing -- original draft, Writing -- review \& editing. \textbf{D.X.}: Formal analysis, Writing -- review \& editing. All authors read and approved the final manuscript.

\subsection*{Institutional Disclaimer}
This work was completed before the first author's affiliation with Heriot-Watt University and was conducted entirely independently of the institution.



\bibliography{sn-bibliography}

\end{document}